\documentclass[11pt, a4paper, logo, nonumbering]{aiwaves}
\usepackage[authoryear, sort&compress, round]{natbib}
\usepackage{dblfloatfix}
\usepackage{ulem}
\usepackage{caption}
\usepackage{dramatist}
\usepackage{xspace}
\usepackage{pifont}
\usepackage{multirow}
\usepackage{tcolorbox}
\usepackage{microtype}
\usepackage{xltabular}
\usepackage{longtable}
\usepackage[hidelinks]{hyperref}

\hypersetup{
  colorlinks=false,    
  linkcolor=black,     
  citecolor=black,     
  urlcolor=black       
}

\usepackage{amsfonts}
\usepackage{amsmath}
\usepackage{amssymb}
\usepackage{lineno}
\usepackage{wrapfig}

\usepackage[bottom]{footmisc}

\usepackage{CJKutf8}
\usepackage{subfigure}
\usepackage{setspace}

\makeatletter
\def\@BTrule[#1]{%
  \ifx\longtable\undefined
    \let\@BTswitch\@BTnormal
  \else\ifx\hline\LT@hline
    \nobreak
    \let\@BTswitch\@BLTrule
  \else
     \let\@BTswitch\@BTnormal
  \fi\fi
  \global\@thisrulewidth=#1\relax
  \ifnum\@thisruleclass=\tw@\vskip\@aboverulesep\else
  \ifnum\@lastruleclass=\z@\vskip\@aboverulesep\else
  \ifnum\@lastruleclass=\@ne\vskip\doublerulesep\fi\fi\fi
  \@BTswitch}
\makeatother

\addto\extrasenglish{
}

 {\begin{list}{}%
         {\setlength{\leftmargin}{#1}}%
         \item[]%
 }
 {\end{list}}

\newcommand{\baby}{\textls[30]{\textsc{Weaver}}\xspace}
\newcommand{\wawawriter}{\textls[50]{\textsc{WawaWriter}}\xspace}
\newcommand{\bench}{\textls[30]{\textsc{WriteBench}}\xspace}

\newcommand{\ultra}{\textls[30]{\textsc{Weaver Ultra}}\xspace}
\newcommand{\pro}{\textls[30]{\textsc{Weaver Pro}}\xspace}
\newcommand{\base}{\textls[30]{\textsc{Weaver Base}}\xspace}
\newcommand{\mini}{\textls[30]{\textsc{Weaver Mini}}\xspace}

\usepackage{todonotes}

\usepackage{appendix}
\usepackage{titlesec}

\reportnumber{001} 

\renewcommand{\today}{}

\title{\centering \baby: 
Foundation Models for Creative Writing}
\author{
    \textbf{Tiannan Wang \ \ Jiamin Chen \ \ Qingrui Jia \ \ Shuai Wang \ \  Ruoyu Fang \ \ Huilin Wang } \\
    \textbf{Zhaowei Gao \ \ Chunzhao Xie \ \  \ \ Chuou Xu \ \ Jihong Dai  \ \ Yibin Liu \ \ Jialong Wu \ \ Shengwei Ding} \\
    \textbf{Long Li \ \  Zhiwei Huang \ \ Xinle Deng \ \ Teng Yu \ \ Gangan Ma \ \ Han Xiao \ \ Zixin Chen } \\
    \textbf{Danjun Xiang \ \ Yunxia Wang \ \ Yuanyuan Zhu \ \ Yi Xiao \ \ Jing Wang \ \ Yiru Wang \ \ Siran Ding } \\ 
    \textbf{Jiayang Huang \ \ Jiayi Xu \ \ Yilihamu Tayier \ \ Zhenyu Hu \ \ Yuan Gao \ \ Chengfeng Zheng} \\
    \textbf{Yueshu Ye \ \ Yihang Li \ \ Lei Wan \ \ Xinyue Jiang \ \ Yujie Wang \ \ Siyu Cheng \ \ Zhule Song} \\
    \textbf{ Xiangru Tang \ \ \ Xiaohua Xu \ \ \  Ningyu Zhang \ \ \ Huajun Chen} \\   
    \textbf{Yuchen Eleanor Jiang\textsuperscript{*} \ \ Wangchunshu Zhou\textsuperscript{*}}
}

\affil{AIWaves Inc.}
\correspondingauthor{Corresponding authors: \{eleanor,chunshu\}@aiwaves.cn}

\renewcommand{\phi}{\varphi}

\renewcommand{\epsilon}{\varepsilon}
\renewcommand{\imath}{\mathrm{i}}

\newlength{\restsubwidth}
\newlength{\restsubheight}
\newlength{\restsubmoreheight}
\newcommand{\rest}[2]{%
        \settowidth{\restsubwidth}{\ensuremath{#2}}
        \settoheight{\restsubheight}{\ensuremath{{}_{#2}}}
        \ensuremath{{#1\hskip 0.5pt}_{\vrule\kern2pt\parbox[b][%
        4pt][b]{\the\restsubwidth}{%
                        \ensuremath{{}_{#2}}}}}
        }

\begin{abstract}
This work introduces \baby, our first family of large language models (LLMs) dedicated to content creation. \baby is pre-trained on a carefully selected corpus that focuses on improving the writing capabilities of large language models. We then fine-tune \baby for creative and professional writing purposes and align it to the preference of professional writers using a suit of novel methods for instruction data synthesis and LLM alignment, making it able to produce more human-like texts and follow more diverse instructions for content creation.  The \baby family consists of models of \textls[30]{\textsc{Mini}} (1.8B), \textls[30]{\textsc{Base}} (6B), \textls[30]{\textsc{Pro}} (14B), and \textls[30]{\textsc{Ultra}} (34B) sizes, suitable for different applications and can be dynamically dispatched by a routing agent according to query complexity to balance response quality and computation cost. Evaluation on a carefully curated benchmark for assessing the writing capabilities of LLMs shows \baby models of all sizes outperform generalist LLMs several times larger than them. Notably, our most-capable \ultra model surpasses GPT-4, a state-of-the-art generalist LLM, on various writing scenarios, demonstrating the advantage of training specialized LLMs for writing purposes. Moreover, \baby natively supports retrieval-augmented generation (RAG) and function calling (tool usage). We present various use cases of these abilities on improving AI-assisted writing systems, including integration of external knowledge bases, tools, or APIs, and providing personalized writing assistance. Furthermore, we discuss and summarize a guideline and best practices for pre-training and fine-tuning domain-specific LLMs. \\

\baby is currently accessible at \url{www.wawawriter.com}, our innovative human-AI collaborative writing platform (For the English version of \wawawriter, see \url{www.wawawriter.com/en}). We discuss a few innovations of the platform from the perspective of human-computer interaction to explain how it will revolutionize traditional AI-assisted writing systems. 

\end{abstract}

\begin{document}
\begin{CJK*}{UTF8}{gbsn}
\maketitle

\newpage

\begin{spacing}{0.9}
\tableofcontents
\end{spacing}

\newpage

\section{Introduction}

Large language models (LLMs)~\citep{gpt,gpt2,gpt3,gpt4,chatgpt,claude,bard, gemini, llama, llama2, mistral,yin2023survey, zhao2023survey} based on Transformers~\citep{transformer} have become a prominent pathway to Artificial General Intelligence (AGI). LLMs acquire massive world knowledge by learning to predict the next word on large-scale web corpora. The capabilities of LLMs have been continuously increasing by scaling model sizes, dataset sizes, and computation. After pre-training, LLMs can be aligned to support real-world use cases by supervised fine-tuning~\citep{flan,t0} and preference optimization techniques including reinforcement learning from human feedback (RLHF)~\citep{rlhf,zheng2023secrets,wang2024secrets} and direct preference optimization (DPO)~\citep{dpo}. The capabilities of LLMs have empowered various applications including ChatGPT, Claude, Bard, Microsoft Copilot, Character.AI, Notion AI, etc. Recently, many specialized LLMs have been trained for different targeted usage scenarios. In general, LLMs specialize according to the targeted domains (e.g., finance~\citep{bloomberggpt}, healthcare~\citep{yang2022large}, legal~\citep{cui2023chatlaw}, etc.) and tasks (e.g., role-playing~\citep{wang2023rolellm}, coding~\citep{codellama}, etc.). However, the ability of LLMs to write human-like texts and produce creative content, which is a critical use case of LLM applications such as ChatGPT, is mostly overlooked by the community.

In this report, we focus on the \textit{\textbf{literature}} domain and the task of \textit{\textbf{writing}}, or \textit{\textbf{content creation}}, and introduce \baby, a family of LLMs dedicatedly pre-trained and aligned for this purpose. The name "Weaver" symbolizes the model's proficiency in skillfully amalgamating linguistic elements, akin to the way a craftsman weaves threads to form a fabric. We answer four main questions in this technical report: \textit{\textbf{why we need \baby}}, \textit{\textbf{how we train \baby}}, \textit{\textbf{how \baby performs}}, and \textit{\textbf{what we build with \baby}}.

\begin{wrapfigure}{r}{8cm} 
    \includegraphics[width=8cm]{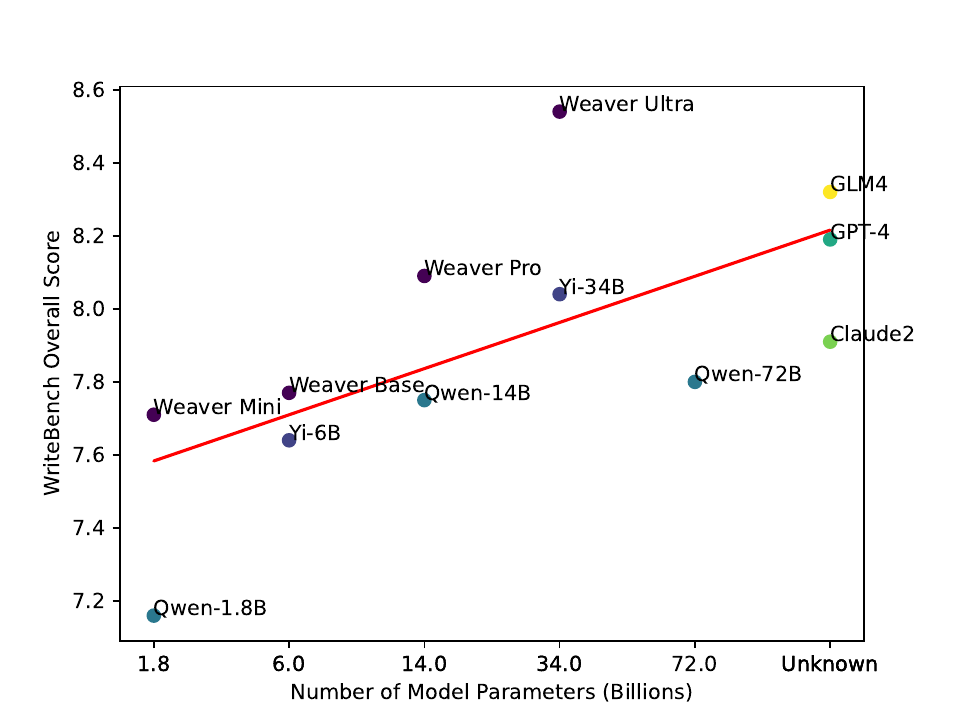}
    \caption{Comparison between \baby and generalist LLMs on \bench.}
    \label{fig:numvsscore}
\end{wrapfigure}

\noindent \textbf{Why we need \baby?}
Despite generalist LLMs such as GPTs already possessing general writing skills and helping billions of users in various writing scenarios, they often struggle to produce human-like texts in specific writing scenarios such as writing stories, fiction, social media copies, blogs, papers/thesis, etc. We analyze the behavior of pre-trained base LLMs such as LLaMA and aligned LLMs such as ChatGPT and LLaMA-chat and believe this limitation originates from both the pre-training stage and the alignment stage. On one hand, generalist LLMs are pre-trained on massive low-quality web texts or machine/AI-generated texts. Consequently, existing LLM backbones tend to produce seemingly fluent texts that are not creative enough and lack human-like styles. On the other hand, during the alignment stage, state-of-the-art LLMs such as GPT-4 are instruction-tuned using instruction-response pairs annotated by crowdsource annotators \citep{wang2023aligning,shen2023large,ji2023ai}. However, most of the annotators are not professional writers or content creators and the annotation guidelines only require them to produce helpful and harmless responses~\citep{instructgpt}. As a result, the crowdsourced data for supervised fine-tuning is less stylish and lacks creativity. Furthermore, most popular preference optimization methods such as RLHF and DPO optimize the model on model-generated data pairs, making them less suitable for enhancing the creativity of LLMs. 

These factors make current generalist LLMs lack creativity and unable to produce human-style texts despite they are super powerful in other applications such as writing codes and answering general questions. We believe this phenomenon will continue to be amplified given that the amount of LLM-generated texts on the internet is exponentially growing and most LLMs are aligned using texts produced by other LLMs. Therefore, we believe it is necessary to train domain-specific LLMs dedicated to writing purposes that are creative and generate human-like texts in order to fully exploit the potential of AI-generated content (AIGC).

\noindent \textbf{How we train \baby?}
To address the aforementioned issues limiting generalist LLMs' creative writing ability, we carefully design a suite of strategies for automated data collection, data annotation, and data filtering for pre-training and alignment. This makes us able to pre-train and align \baby on diverse, human-like, and stylish texts.  To be specific, we conduct extensive pre-training data filtering and only keep high-quality content such as books, fiction, stories, and articles in the pre-training corpus, making the pre-trained backbones more likely to produce human-like texts.

As for the alignment stage, we propose a new instruction backtranslation framework inspired by LongForm~\citep{longform} and Humpback~\citep{humpback} that synthesize diverse and natural instructions that correspond to high-quality outputs written by professional writers and preferred by human consumers. Our instruction backtranslation framework translated the work of crowdsource annotators from writing both instructions and outputs to simply collecting high-quality content such as stories, fiction, articles, social media copies, and blog posts. This massively reduces the cost of instruction data annotation and the requirement for crowdsource annotators while significantly improving the quality of annotated data.

Moreover, we also propose a novel Constitutional DPO algorithm for preference optimization to better align \baby to the preference of professional writers and content creators. Constitutional DPO is inspired by and combines the advantages of a few previous works including DPO~\citep{dpo}, Constitutional AI~\citep{constitutionalai}, Self-Align~\citep{salmon}, and RLCD~\citep{rlcd}. Specifically, Constitutional DPO exploits expert (e.g., professional editors in our case) annotated principles to synthesize negative examples that violate certain principles based on positive examples that are sampled from the optimal policy (e.g., texts produced by professional writers or content creators in our case). In contrast to the common practice of using DPO that uses LLMs to produce preference annotation on two model-generated responses such as Zephyr~\citep{zephyr}, the pairwise preference data synthesized by our approach contains less noise since the negative example are deliberately synthesized to be of lower quality compared to the positive example. The pairwise preference data generated by Consitutional DPO also contains more principled and targeted learning signals that can be adjusted by human experts according to target domains and applications.

Furthermore, we propose to transform the annotation instructions and responses used in the instruction backtranslation and Constitutional DPO stages into annotation instructions and evaluation instructions. In this way, \baby not only possesses abilities to follow writing instructions but can also annotate writing instructions and evaluate writing outputs. We also curate instruction data for retrieval-augmented generation (RAG) and function calling to enable \baby to exploit external knowledge and tools. The combination of different data sources makes \baby a versatile foundation model while specializing in creative writing.

\noindent \textbf{How \baby performs?}
Evaluating the content creation/writing ability of LLMs remains an open problem since existing benchmarks for LLMs such as MMLU~\citep{mmlu} or MT-Bench~\citep{mtbench} mostly focus on reasoning, math, coding, or general questions instead of creative writing. Moreover, it is already notoriously hard to evaluate LLMs on general instructions, and it becomes much harder for creative writing tasks since literary critic is non-trivial even for human experts, not to mention LLMs. To better evaluate \baby and help the LLM community better measure progress on AIGC, we carefully curate \bench, a benchmark for assessing the creative writing capabilities of LLMs and collect outputs from 10$+$ popular LLMs covering both open-source and proprietary models.

We then conduct both LLM-based and human evaluation of \baby and generalist LLMs on the benchmark. Evaluation results confirm the superiority of \baby compared to generalist LLMs. We find that \ultra, the most-capable model in the \baby family, advances the state-of-the-art in creative writing despite being 10$+$ smaller compared to GPT-4\footnote{According to non-official rumor about the size of GPT-4}, the previous best performing LLM. Other models in the \baby family also surpass competitive generalist LLMs several times larger than them. Our analysis and case studies show that the main source of improvements is because \baby can generate texts that are creative and human-like while generalist LLMs tend to produce too ``predictable'' texts. To confirm that \baby is \textit{truly helpful} in real-world applications, we also conduct a user study where human writers are asked to write stories (fiction writing) and blog posts (non-fiction writing) with \baby and GPT-4. Our user study shows that compared to GPT-4, \baby improves the writers' productivity by 47\% and helps writer produce better stories and articles at the same time.


\noindent \textbf{What we build with \baby?}
Training specialized LLMs for writing is one side of enhancing AI-assisted writing experience. We believe it is also very important to build a better human-AI interface to fully exploit the potential of \baby on AI-assisted writing. To this end, we introduce \wawawriter, our innovative human-AI collaborative writing
platform. Similar to recent AI writing products such as Notion AI, \wawawriter provides a chat interface that allows users to provide diverse writing instructions, instead of merely suggesting the next one or few sentences based on the current context or polishing the content as in traditional applications. \wawawriter also takes a few steps further: (1) we enable \textbf{\textit{human-AI co-editing}} by allowing users to customize language agents~\citep{agents} that acts like a human collaborator by operating inside the editor simultaneously with users; (2) we allow users to build \textbf{\textit{personal knowledge bases}} by saving websites or uploading documents and build a RAG pipeline that integrates knowledge bases to \baby; (3) we propose to provide \textbf{\textit{personalized writing assistance}} by analyzing users' personal writing styles using LLMs based on their writing history on the platform and using the results to guide \baby's text generation process. By combining these innovations, \wawawriter aims to provide next-generation AI-assisted writing experience that is more helpful and enjoyable. 

In the following sections, we first describe the architectures and sizes of the \baby family and their pre-training stage. We then present details on the abilities of \baby, how we synthesize training data to help \baby acquire these abilities and learn to produce human-like stylish texts, and the details for the alignment stage. We also present our benchmark for evaluating the writing abilities of LLMs and the evaluation results. Finally, we introduce the details of \wawawriter and present how \baby paves the way for next-generation AI-assisted writing experiences. 
\section{Pre-training}

\begin{table}[hb]
    \centering
    \begin{tabular}{ccccccccc}
    \toprule
        \multirow{2}{*}{Name} & \multirow{2}{*}{Params} & \multirow{2}{*}{$n_{\mathrm{layers}}$} & \multirow{2}{*}{$d_{\mathrm{model}}$} & \multirow{2}{*}{$n_{\mathrm{heads}}$} & Context & Sequence & Learning & \multirow{2}{*}{Tokens} \\
         & & & & & Length & Batch Size & Rate & \\
    \midrule
        \mini & 1.8B  & 24 & 2048 & 16 & 4096 & 512 & 1e-4 & 50B \\
        \base & 6B  & 32 & 4096 & 32  & 4096 & 512 & 1e-4 & 50B \\
        \pro & 14B  & 40 & 5120 & 40 & 4096 & 512 & 1e-4 & 40B \\
        \ultra & 34B & 60 & 7168 & 56   & 4096 & 520 & 5e-5 & 18B \\
    \bottomrule
    \end{tabular}
    \caption{Description for the \baby family.}
    \label{tab:architecture}
\end{table}

\subsection{Model Family}

\baby models are language models built on top of Transformer decoders. We have adopted the recent improvements from the design of LLaMA~\citep{llama,llama2}, the most popular open-source LLM, including a Pre-Norm structure with RMSNorm~\citep{rmsnorm} function, SwiGLU~\citep{glu} as the activation function for the Feed-Forward Network, Rotary Embedding \citep{rope} for positional encoding, and Grouped-Query Attention (GQA) \citep{gqa}.

The \baby family consists of models of four different sizes: \textls[30]{\textsc{Mini}},  \textls[30]{\textsc{Base}}, \textls[30]{\textsc{Pro}}, and \textls[30]{\textsc{Ultra}}, ranging from 1.8B to 34B parameters. We train different model sizes to support different applications as the complexity of writing tasks varies a lot across different domains and use cases. All \baby models are initialized from powerful open-source LLMs. We provide detailed configurations and descriptions of \baby models in Table \ref{tab:architecture}. 

\subsection{Pre-training Data}
We then present an overview of pre-training data selection strategies and the resulting pre-training data mixture. Since \baby models are initialized from powerful open-source LLMs and thus already possess adequate world knowledge, the amount of continual pre-training data does not need to be super large. We consider the continual pre-training stage to be the process where \baby learns to reallocate or re-balance its capabilities: the model allocates more capabilities to writing and content creation while reducing the capabilities on other domains such as mathematics and coding. 

Therefore, we only include manually verified data sources including various kinds of content such as books, fiction, stories, news articles, papers, reports, social media copies, etc., in the pre-training data. We combine rule-based and machine-learning-based methods to filter low-quality texts. In addition to data sources and filtering, we also carefully control the data mixture between different domains. Specifically, we mix fiction data (i.e., fiction and stories) and non-fiction data (i.e., articles, papers, reports, etc.) with a ratio of $1:1$. We also mix Chinese and English data with a portion of $4:1$ to make \baby supports both Chinese and English.

\subsection{Training Details}
We train \baby using the standard autoregressive language modeling task where the model learns to predict the next token based on the context of previous tokens.
We train \baby models with a context length of 4096. We shuffle and merge the documents, and then truncate them to the specified context lengths to create training batches.
We incorporate Megatron-Deepspeed~\citep{megatron} and Flash Attention2~\citep{flashattention,flashattention2} to improve computational efficiency and reduce memory usage.
We adopt the standard optimizer AdamW~\citep{adamw} and set the hyperparameters $\beta_1=0.9$, $\beta_2=0.95$, and $\epsilon=10^{-8}$. We use a cosine learning rate schedule with a specified peak learning rate for each model. The learning rate is decayed to a minimum learning rate of $10\%$ of the peak learning rate. All models are trained with BFloat16 mixed precision for training stability. We present detailed pre-training configurations for each model in Table \ref{tab:architecture}.

\section{Data Synthesis}
After pre-training, \baby models contain a large amount of world knowledge and writing skills and can produce human-like texts conditioning on high-quality contexts. To unlock these capabilities for real-world applications, we need to curate a high-quality dataset for alignment. The format and quality of the dataset significantly affect the coverage of abilities and the quality of aligned models. As discussed in the Introduction, the common practice for alignment data collection of existing generalist LLMs severely limits their writing capabilities. In this section, we describe our data synthesis framework in detail. We first describe the abilities we want to unlock during the alignment stage and then present our proposed data synthesis methods for both the supervised fine-tuning and the preference optimization stage.

\subsection{Abilities}
We first describe the categories of abilities we want to unlock for \baby during the alignment stage.
\subsubsection{Instruction Following}
The first obvious ability we need to unlock is the ability to follow writing instructions and produce human-like stylish texts. We cover various domains and tasks as listed below during data collection and alignment training. 

\paragraph{Domains}
\ \\
\\
\noindent\textbf{Fiction Writing:}
Fiction writing refers to the abilities of models to write stories and fiction. We divide fiction writing into several subdomains with respect to the length and the genre of the fiction. We cover fiction and stories of lengths ranging from a few hundred to a few million characters, and fiction types including sci-fiction, romance, fantasy, horror, mystery, and thriller. 

\noindent\textbf{Creative Non-Fiction Writing:}
Creative non-fiction writing is a genre of writing that uses literary styles and techniques to create factually accurate narratives. We cover typical creative non-fiction writing cases including writing memoirs, biography, travelogue, journalism, social media copy, blog posts, news articles, commentary, etc. 

\noindent\textbf{Marketing Writing:}
We also consider marketing writing, which involves writing business plans, advertising copies, product promoting, marketing plans, etc. Marketing writing differs from previous categories because it is highly application-oriented and the style of generated texts is not the most important. However, marketing writing still requires human-like creativity to attract potential users.

\noindent\textbf{Technical Writing:}
Technical writing includes tasks such as paper writing, patent writing, report writing, etc. Technical writing requires more accuracy compared to creativity. However, writing-specific training can still be helpful because it can help model produce texts that accurately adhere to the style required for specific scenarios.

\paragraph{Tasks}
\ \\
\\
\noindent\textbf{Content writing:}
Content writing is the basic task that requires the model to generate content (i.e., fiction, articles, etc.) based on certain instructions. Writing instructions vary in terms of whether the previous context is provided and how fine-grained the given instructions are. The task requires the LLM to be able to understand and adhere to specific requirements expressed in the instructions while also producing texts that are consistent and coherent with previous contexts. For example, a typical content writing instruction is: ``Please help me write a sci-fi about what will happen after people finally achieve AGI.''

\noindent\textbf{Outlining:}
Outlining is the task of writing outlines, which is a common practice for writers in both fiction and non-fiction writing. As discussed in the literature of long text generation~\citep{hierarchial_gen,sun-etal-2022-summarize, re3, doc,recurrentgpt}, it is often helpful to let the model first generate an outline before generating long texts. Outlines vary according to different domains and the granularity/length of outlines. One example for the task of outlining is ``Please help me write an outline of my annual work report.''

\noindent\textbf{Polishing \& Editing:}
Polishing and editing require the model to improve the quality of a paragraph or rewrite it following the requirements expressed in the instructions. The task is closely related to the task of grammatical error correction~\citep{conll14gec,bea19gec} with a key difference that the modifications are not necessarily grammatical errors. Compared to the task of academic writing polishing described in~\citet{doolittle}, we support customized fine-grained control of polishing or editing requirements, which is important for human-AI interaction in AI-assisted writing systems. A typical polishing instruction may look like this: ``Please help me revise the following texts, keep in mind that the revised texts should be suitable for an academic paper.''

\noindent\textbf{Style Transferring:}
The task of text style transfering requires the model to transform texts in one style into another style. For example, one may want to transform a story into a script or turn a report into a speechwriting. We cover both template-based style transfer that uses a template to provide target style information~\citep{rag,guu-etal-2018-generating} and description-based style transfer which uses either a keyword~\citep{ctg} or a short description~\citep{instructctg} for the target style. For example, one may ask the model to ``Transform the following book chapter into a script.''

\noindent\textbf{Expanding/Simplifying:}
Text expanding and simplifying requires the model to revise an input paragraph to make it longer or shorter according to certain instructions. Text summarization and summary-to-article generation can be regarded as two extreme cases of this task. One exemplar instruction is: ``Please help me summarize this paragraph into one sentence.''. 

\noindent\textbf{Brainstorming:}
Brainstorming requires the model to help users come up with creative ideas based on the current context and user instructions. A typical brainstorming instruction is: ``Please give me 5 possible character descriptions for a villain to appear in the next chapter, including his name, appearance, occupation, and background.''

\noindent\textbf{Reviewing:}
Reviewing refers to the task of reading and analyzing a given piece of text critically and then producing comments or revising suggestions. For example, one may ask the model to ``Please take a look at my essay and list 5 suggestions to improve it.''

\subsubsection{Instruction Annotation}
We also train \baby to support the instruction annotation task. As described in Humpback~\citep{humpback} and LongForm~\citep{longform}, given a piece of text, the task requires the model to generate an instruction to which the input texts may be the answer. However, vanilla instruction backtranslation only supports the writing task. Therefore, for instruction annotation, we require the model to synthesize an instruction-response pair based on a text span. The response can be the text span, a part of the text span, or inferred from the text span. This substantially broadens the scope for vanilla instruction backtranslation since most automatically mined text spans may not be suitable for a certain instruction on itself while a part of the text span can be a valid response or one may construct a high-quality instruction-response pair based on it. The instruction annotation ability enables \baby to mine training data for itself on large-scale corpus, opening the possibility of \textit{scalable self-training on web data}.

\subsubsection{Evaluation (Literary Critic)}
Many recent work explored using or training LLMs to evaluate general instruction following tasks~\citep{pandalm,tigerscore,chateval}.
However, we find generalist LLMs require extensive prompting skills to make them suitable for evaluating tasks related to creative writing. Moreover, since almost all students majoring in creative writing are also required to take literary critic courses, we think learning to perform literary critic may be helpful for the model to produce better texts as well. Therefore, we also train \baby to judge the quality of the responses to writing instructions and do pairwise comparison of two responses.

We collect human preference between model outputs in \wawawriter, our AI-assisted writing platform and convert the collected preference data to training data for LLM-based evaluation with carefully curated templates.

\subsubsection{Retrieval-Augmented Generation}
The ability of retrieval-augmented generation (RAG)~\citep{rag,gao2023retrieval}, i.e., generating responses by referring to external knowledge or references as context. RAG is an important technique that helps LLMs generate more accurate and informed responses. It can be especially helpful for writing purposes since it's common for human writers to refer to other text samples when writing fiction or articles. However, most existing LLMs purely rely on prompt engineering to do RAG and do not perform RAG training during alignment. We believe this limits the ability of LLMs to make use of retrieved contexts. Therefore, we propose to include RAG-aware training data during alignment to enhance \baby's retrieval-augmented generation ability. Specifically, we augment 10\% percent of training data by appending a relevant context obtained by retrieving the paragraph most similar to the target response. In this way, \baby learns to write by referring to external contexts and is thus more compatible with RAG techniques compared to most existing LLMs.  

\subsubsection{Function Calling}
The ability to use tools is also very important for LLMs~\citep{toolformer}. This ability, also referred to as ``function calling'', is also helpful for writing because the model may need to search the internet for references or call editor APIs when doing human-AI collaborative writing. To unlock the function calling ability, we include an open-source function calling dataset\footnote{https://huggingface.co/glaiveai} into supervised fine-tuning data. We also propose a new pipeline to synthesize more diverse function calling data by first using GPT-4 to synthesize diverse environments with multiple tools and APIs, as well as their documentation. We then randomly select one API at a time and ask GPT-4 to imagine a situation where the API can be helpful and the plausible arguments for the API. We then reason what one may instruct an LLM in that situation so that the API should be used with the arguments. Finally, similar to how GPTs support function calling, we train \baby to use tools by selecting the right API and generating the arguments given the instructions and the contexts.

\begin{table}[!h]
\centering
\begin{tabularx}{\textwidth}{llXXl}
\toprule
\textbf{Domain} & \textbf{Subdomain} & \textbf{Description} & \textbf{Source} \\ \midrule
\textbf{Fiction Writing}        &  \textbf{Full Novel}   & Web novel, over 1M words      & Proprietary     \\ \midrule
                & \textbf{Short Story}        & Web stories, 10k-20k words      & Proprietary      \\ \midrule
\small \textbf{Creative Non-Fiction Writing}      & \textbf{Red}      & Top liked and commented posts on Red      & Picked \\ \midrule     
                & \textbf{Zhihu}          & Top upvoted posts on Zhihu      & Picked \\ \midrule
                & \textbf{Weibo}         & Top liked posts on Weibo      & Picked      \\ \midrule
                & \textbf{WeChat Articles}          & Top read articles on WeChat       & Picked      \\ \midrule
                & \textbf{DouBan}          & Top liked posts on DouBan        &  Picked   \\ \midrule   
                & \textbf{News \& Blogs}          & Popular news/blogs    & Picked      \\ \midrule    
\textbf{Technical Writing}         & \textbf{Papers}        & Academic papers on CNKI   &  Picked  \\ \midrule
                & \textbf{Essay}       & Online essays     & Picked    \\ \midrule
                & \textbf{Contract}      &  Contracts from online sources     & Picked      \\ \midrule
                & \textbf{Reports}      & Reports for work, business, science, etc.      & Proprietary   \\ \midrule
                & \textbf{Copies}        & Business \& Government copies      & Proprietary      \\ \midrule

\textbf{Marketing Writing}          &   \textbf{Business Plans}         & Business plans for projects and startups      &  Proprietary  \\ \midrule
                & \textbf{Industry Report}         & Research report for different industries     & Proprietary  \\ \midrule
                & \textbf{Advertising Copy}          & Popular copies for advertisements     & Picked      \\ \midrule
                & \textbf{Marketing Plan}          & Marketing plans for products \& services      & Picked      \\ \midrule
                & \textbf{Product Overview}       & Articles advertising products      & Picked     \\ \bottomrule
\end{tabularx}
\caption{Description of SFT Data sources. We combine similar subdomains in the same fields for simplicity. The entire training set covers 34 subdomains and around 500,000 instruction-output pairs. ``Picked'' means the raw data in the corresponding domains are manually selected.}
\label{tab:sft_data_sources}
\end{table}

\subsection{Instruction Backtranslation}
We then describe our proposed improved pipeline for instruction backtranslation. The motivation for doing instruction backtranslation instead of instruction augmentation methods such as self-instruct~\citep{self-instruct} is very simple: we want to align \baby on high-quality, stylish, and human-written texts. To achieve this goal, we first collect high-quality stories, fiction chapters, and copies of different domains. We list the categories of collected texts in Table \ref{tab:sft_data_sources}. 

We then use a carefully designed few-shot prompt template to synthesize instruction-response pairs for all aforementioned writing tasks. Specifically, for each subdomain-task pair, we annotate 5 cases of how one can write an instruction-response pair, including both the annotated results and the rationales for the annotation process: we first select a text span from a case as the output (except for outlining, brainstorming, and reviewing tasks where the output is transformed from the selected text span with an additional prompt). We then identify or produce the context for the output. For example, for the polishing task, the context should be a worse version of the target output, so we can modify the wording and structure of the target output to make it look worse. Then we infer the instruction that one may use to transform the context to the output. Taking the polishing task as an example again, we need to reason what modifications are made and synthesize the polishing instructions accordingly. For each unlabeled case, we use the annotated cases as few-shot exemplars and ask GPT-4 to first generate the annotation process in the Chain-of-Thought style~\citep{cot} and then produce the synthesized instruction-response pairs. The instruction backtranslation pipeline is illustrated in Figure 1. We synthesize 500,000 high-quality instruction-response pairs across all domains and tasks with this pipeline. Finally, we do an instruction data selection procedure following the practice described in~\citep{liu2023makes}: we first score all instruction-response pairs with GPT-3.5-turbo and then select top-ranked data in each subdomain-task pair for supervised fine-tuning. Specifically, we score each instruction-response pair based on the quality and the diversity of the instruction and the relevance between the instruction and the response.

\begin{figure}[htbp]
\centering
\includegraphics[width=\textwidth]{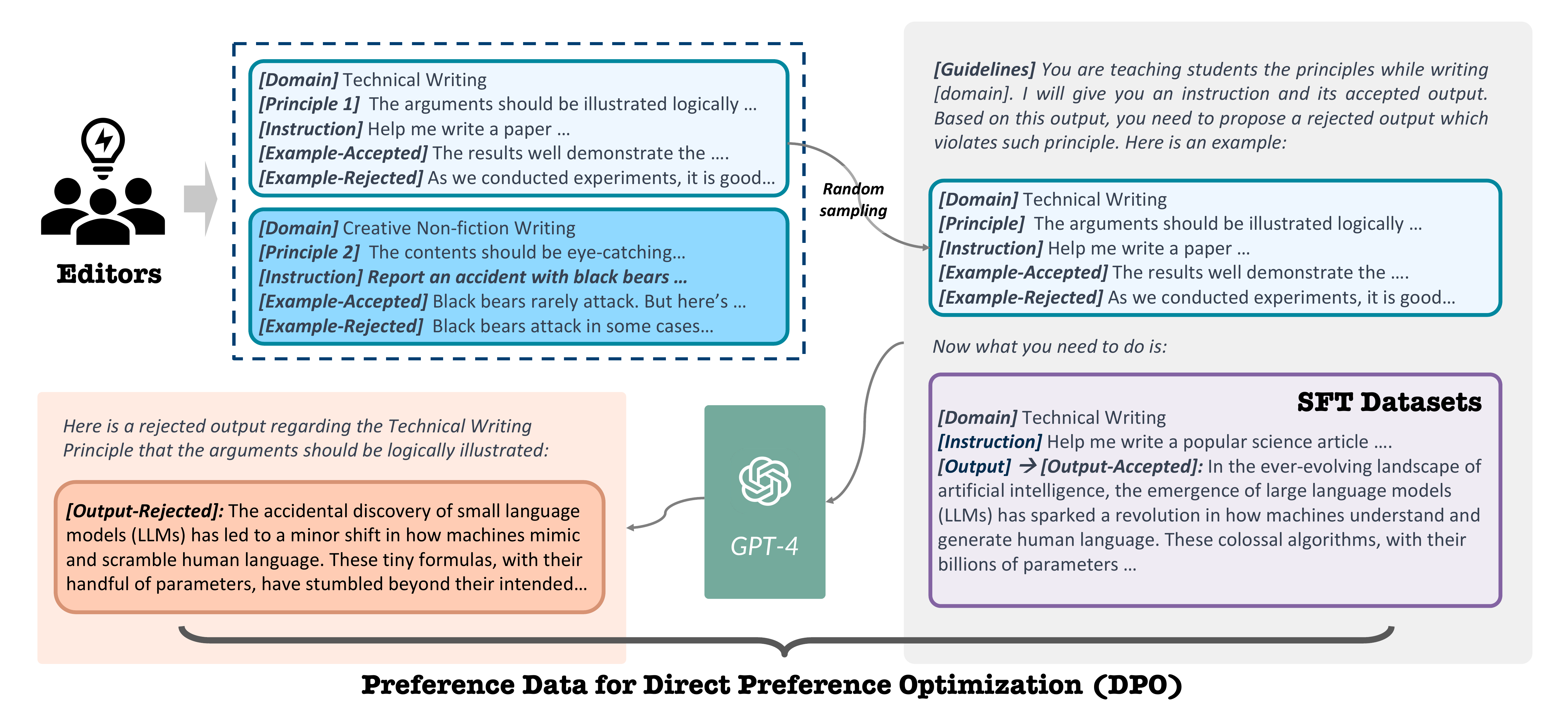} 
\caption{Illustration of the Constitutional DPO framework.} 
\label{fig:constitutionaldpo}
\end{figure}

\begin{table}[ht]
\centering
\caption{Examples of expert-annotated principles in four domains and sampled tasks.}
\begin{tabular}{@{}lll@{}}
\toprule
Domain                                                                                   & Task                 & Principles                                                                                                                                      \\ \midrule
\multirow{3}{*}{\begin{tabular}[c]{@{}l@{}}Creative Non-\\ fiction Writing\end{tabular}} & Content Writing      & \begin{tabular}[c]{@{}l@{}}The content should be created to encourage readers to \\ engage in interactions, comments, etc.\end{tabular}         \\
                                                                                         & Polishing \& Editing & The revised content should align with the original text.                                                                                        \\
                                                                                         & Brainstorming        & The content should refrain from pre-judging ideas.                                                                                              \\ \midrule
\multirow{2}{*}{Technical Writing}                                                       & Content Writing      & \begin{tabular}[c]{@{}l@{}}The generated content should avoid bias toward certain \\ genders, professions, regions, etc.\end{tabular}           \\
                                                                                         & Style Transferring   & \begin{tabular}[c]{@{}l@{}}The style of the content should be consistent with the \\ language style specified in the instructions.\end{tabular} \\ \midrule
\multirow{2}{*}{Fiction}                                                                 & Content Writing      & \begin{tabular}[c]{@{}l@{}}The perspective should remain consistent with the out-\\ line or previous content.\end{tabular}                      \\
                                                                                         & Outlining            & \begin{tabular}[c]{@{}l@{}}The global outline should not be too brief or general, \\ omitting key plot points.\end{tabular}                     \\ \midrule
\multirow{2}{*}{Marketing Writng}                                                        & Content Writing      & The content of the market writing should be accurate.                                                                                           \\
                                                                                         & Summarizing          & \begin{tabular}[c]{@{}l@{}}The summarized content should be all-encompassing,\\ leaving out no crucial points.\end{tabular}                     \\ \bottomrule
\end{tabular}
\end{table}

\subsection{Constitutional DPO: Learning From Principled Negative Examples}
Finally, we propose Constitutional DPO, a novel alignment method that encourages LLMs to learn from preference data consisting of samples from the optimal policy and ``principled'' negative examples synthesized with AI feedback. Our approach combines the advantages of Constitutional AI~\citep{constitutionalai,salmon}, which train reward models based on principles written by human experts, RLCD~\citep{rlcd}, which prompt LLMs to generate positive/negative examples and train reward models with AI-generated preference data, and DPO~\citep{dpo}, which omits reward model training and does direct preference optimization.

Specifically, we first invite human experts including professional writers, editors, and content creators to annotate principles for different writing tasks. Different from previous ``principle-based'' approaches that only write a short description of the principles, for each principle we also collect one case adhering to the principle and one case violating the principle, as well as natural language rationales explaining why the cases adhere or violate the principle. Then we sample a subset of the instruction data with the highest scores in the aforementioned data filtering process and consider them as samples from the optimal policy as the output texts are carefully selected and instruction-output pairs are top-ranked. For each sample, we first present the principles for the task and ask GPT to analyze which principle can best explain why the response is of good quality. We then ask GPT to synthesize a counterpart of the response violating the principle while adding minimal modifications and do not affect other good aspects of the original response. 

With the collected data, we consider the original-perturbed response pairs as $(y_w,y_l)$ pairs and do standard DPO training. In this way, each data pair contains critical training signals about the corresponding principles and helps fine-tune the model to follow the principles. The preference data synthesized by our approach contains much less noise compared to standard RLAIF pipeline, especially in writing domains since LLMs struggles to do literary critic. Compared to RLCD, the most related method for preference data generation, we consider high-quality SFT data instead of LLM-generated as positive examples and use expert-written principles for negative example generation. This makes the training signal less noisy and more principled.

\section{Alignment}

\subsection{Supervised Fine-tuning}

\subsubsection{Data}
To collect the dataset for supervised fine-tuning, we first collect high-quality content written by human writers and content creators according to their metadata including their ratings, number of reads, upvotes, and comments. We adopt the aforementioned data synthesis framework to synthesize instruction following data covering 30+ fine-grained domains and over 10 tasks, instruction annotation data, text generation evaluation data, retrieval-augmented generation data, and function calling data. The combined instruction tuning dataset consists of around 1,000,000 samples. We then run the data filtering process and select 400,000 data points as the final dataset for supervised fine-tuning.

\subsubsection{Training}

We fine-tune the continual pre-trained models for 3 to 5 epochs. We use a cosine learning rate scheduler with a peak learning rate of 1e-5 and 2e-5 for larger models (i.e., \ultra and \pro) and 4e-5 for smaller models (i.e., \base and \mini) with 5\% warmup steps. We train all models with a global batch size of 256. After supervised fine-tuning, we select the best-performing checkpoint on an internal validation set for preference optimization.

\subsection{Preference Optimization}

\subsubsection{Data}
For preference optimization, we select 500 highest-rated samples in the data filtering stage for each subdomain as positive examples for the Constitutional DPO pipeline. We collect over 200 principles and their corresponding few-shot exemplars. We generate one negative example per positive example, resulting in 25,000 preference data pairs.

\subsubsection{Training}
We fine-tune the supervised fine-tuned models using the conventional DPO algorithm. We train our models for three to five epochs. We use a linear learning rate scheduler with a peak learning rate of 5e-7 and 5\% warmup steps. We train \ultra using a global batch size of 40, while for the others we use 32 and set $\beta$ = 0.1. We select the best-performing checkpoint on the internal validation set as the final \baby models.

\section{Evaluation}

\subsection{\bench}
Most existing benchmarks for LLMs~\citep{mtbench} and natural language generation~\citep{lin-etal-2020-commongen, jiang-etal-2023-discourse} focus on the reasoning ability or the general-purpose instruction following ability instead of the ability of LLMs to produce creative, stylish, and human-like text content. To this end, we construct \bench, a new benchmark for assessing the writing capabilities of LLMs\footnote{\bench will be publically available at \url{https://github.com/aiwaves-cn/WriteBench}}.

Similar to how we collect training data for \baby, \bench is designed to cover multiple domains and tasks. To ensure a fair comparison between \baby and compared generalist LLMs, the data collection and data selection process for instructions in \bench is done by our independent evaluation team. The resulting \bench consists of over 1000 testing instructions covering four domains including fiction writing, creative non-fiction writing, technical writing, and marketing writing. The first release of the \bench benchmark is in Chinese since we want to measure the Chinese writing capabilities of the compared models.

\subsection{Compared Models}
We compare \baby with competitive Chinese LLMs including both open-sourced models and proprietary models of different sizes, including GPT-4, GPT-3.5, GLM-4, Claude2, Gemini Pro, ERNIE-Bot-4.0, ERNIE-Bot-3.5, Qwen-72B-Chat, Qwen-14B-Chat, Qwen-7B-Chat, Qwen-1.8B-Chat, YI-34B-Chat, YI-6B-Chat, and ChatGLM3-6B. We directly use the same instructions in \bench as input prompts for all tested LLMs and collect the model outputs as responses.

\begin{table}[ht]
\centering
\caption{LLM-based Evaluation Results}
\begin{tabular}{@{}l|cccc@{}}
\toprule
\multicolumn{1}{c|}{\textbf{Models}} & \textbf{Style} & \textbf{Relevance} & \textbf{Creativity} & \textbf{Overall} \\ \midrule
\textbf{\ultra}                   & 8.94           & 8.96               & 7.71                & 8.54             \\
\textbf{GLM-4}                        & 8.83           & 9.55               & 6.58                & 8.32             \\
\textbf{GPT-4}                       & 8.80           & 9.45               & 6.32                & 8.19             \\
\textbf{\pro}                   & 8.52           & 8.45               & 7.3                 & 8.09             \\
\textbf{YI-34B-Chat}                 & 8.70           & 9.17               & 6.26                & 8.04             \\
\textbf{Claude2}                     & 8.42           & 8.89               & 6.41                & 7.91             \\
\textbf{Qwen-72B-Chat}               & 8.47           & 8.98               & 5.95                & 7.80             \\
\textbf{\base}                    & 8.61           & 8.81               & 5.89                & 7.77             \\
\textbf{Qwen-14B-Chat}               & 8.51           & 8.85               & 5.89                & 7.75             \\
\textbf{\mini}                  & 8.41           & 8.38               & 6.35                & 7.71             \\
\textbf{Gemini Pro}                  & 8.39           & 8.79               & 5.88                & 7.69             \\
\textbf{Qwen-7B-Chat}                & 8.40           & 8.80               & 5.81                & 7.67             \\
\textbf{Yi-6B-Chat}                  & 8.24           & 8.67               & 6.00                & 7.64             \\
\textbf{ChatGLM3-6B}                 & 8.16           & 8.70               & 5.86                & 7.57             \\
\textbf{GPT-3.5}          & 8.37           & 8.65               & 5.60                & 7.54             \\
\textbf{ERNIE-Bot-3.5}               & 8.24           & 8.22               & 5.71                & 7.39             \\
\textbf{ERNIE-Bot-4.0}               & 8.15           & 8.05               & 5.61                & 7.27             \\
\textbf{Qwen-1.8B-Chat}              & 7.97           & 7.86               & 5.66                & 7.16             \\ \bottomrule
\end{tabular}
\label{table:llm_eval}
\end{table}
\subsection{LLM-based Evaluation}
We first perform an LLM-based evaluation to do a coarse-grained evaluation of the compared models. We use GPT-4 as the judge to score each instruction-response pair following the practice and prompt templates in MT-Bench. The results are shown in Table~\ref{table:llm_eval}. We find that in terms of writing style and creativity, \ultra significantly outperforms all proprietary models including strong competitors such as GPT-4 and GLM-4. GPT-4 and GLM-4 are better at the relevance metric because they are at least few times larger than \ultra and thus have better instruction-following ability. As for \baby of other sizes, we can see that  with only 14B parameters, \pro outperforms all open-source models including those with 70B and 34B parameters, as well as most proprietary models. Similarly, \base and \mini are also comparable with generalist LLMs with more than two times their sizes. Overall, the results confirm the effectiveness of our data synthesis and training framework for LLMs specialized in creative writing.


\begin{table}[ht]
\centering
\caption{Human Preference on Fiction Writing with the Elo Ranking System}
\label{table:human-fiction}
\begin{tabular}{
  @{}
  l
  c
  c
  c
  c
  c
  @{}
}
\toprule
\textbf{Models} & {\textbf{Creativity}} & {\textbf{Style}} & {\textbf{Relevance}} & {\textbf{Fluency}} & {\textbf{Overall}} \\
\midrule
\textbf{\ultra}         & \bf 1682 & \bf 1661 & \bf 1689 & \bf 1641 & \bf 1657 \\
\textbf{GPT-4}                  & 1507 & 1513 & 1421 & 1534 & 1508 \\
\textbf{ERNIE-Bot-4.0}          & 1404 & 1409 & 1564 & 1544 & 1477 \\
\textbf{Gemini Pro}             & 1513 & 1469 & 1409 & 1360 & 1430 \\
\textbf{GLM-4}                  & 1391 & 1445 & 1415 & 1417 & 1425 \\
\bottomrule
\end{tabular}
\end{table}

\begin{table}[ht]
\centering
\caption{Overall Human Preference with the Elo Ranking System}
\begin{tabular}{
  @{}
  l
  c
  c
  c
  c
  c
  @{}
}\toprule
\textbf{Models}        & \textbf{Creativity} & \textbf{Style}   & \textbf{Relevance} & \textbf{Fluency} & \textbf{Overall} \\ \midrule
\textbf{\ultra}        & \textbf{1589}    & \textbf{1590} & \textbf{1593}   & \textbf{1588} & \textbf{1576} \\
\textbf{GLM-4}         & 1482            & 1527          & 1491   & 1513          & 1521    \\
\textbf{GPT-4} & 1468      & 1505          & 1427            & 1501     & 1501          \\
\textbf{Gemini Pro}    & 1548  & 1490          & 1434 & 1380  & 1454    \\
\textbf{ERNIE-Bot-4.0}         & 1410   & 1385          & 1552  & 1515          & 1445          \\ \bottomrule
\end{tabular}
\label{table:human-all}
\end{table}
\subsection{Human Evaluation}
We then perform a human evaluation to compare \baby with a few representative LLMs including GPT-4, GLM-4, ERNIE-Bot-4.0, and Gemini-pro. We recruit 44 professional Chinese writers or editors as human annotators in human evaluation. We adopt the practice in the ChatBot Arena\footnote{\url{https://chat.lmsys.org/}} benchmark and let human annotators perform three-way pairwise comparisons between two model outputs according to their creativity, stylish, relevance, and fluency. We collect 3540 comparison results and compute the ELO rating of the compared models. The results on fiction writing and the  overall comparison are shown in Table~\ref{table:human-fiction} and Table~\ref{table:human-all}, respectively. We can see that professional writers and editors rates \ultra significantly better than compared models across all metrics. As for other compared models, we find that GPT-4 and Gemini Pro are considered to produce more creative and human-like texts compared to GLM-4 and ERNIE-Bot, we suspect this is because GLM and ERNIE are aligned using GPT distillation data, which probably harms their creativity.

\subsection{User Study}
A good LLM for AI-assisted writing should not only be best-performing on benchmarks but also \textbf{truly helpful} in real-world writing scenarios. To evaluate how truly helpful \baby is, we conduct a user study where 5 professional writers are recruited as subjects. Each subject is provided with two chat interfaces, one with \ultra and the other with GPT-4. We then let each subject write two short stories (with two carefully selected topic) of around 6,000 words with two same chat interfaces powered by GPT-4 and \ultra respectively\footnote{To ensure fair comparison, we give enough time and trials for the writers to get familiar with the interface and the models.}. We measure the time used by the same writer for finishing the two stories and ask a professional editor to judge their quality. We find that compared to GPT-4, \ultra improves the efficiency of the writer by around 3 times. Furthermore, out of 5 topics, the human editor prefer \baby generated story for 4 times and can not decide the winner for the remaining topic. Our user interview reveals that the efficiency improvement mainly comes from the fact that \baby is faster and generates more human-like texts that require less post-editing.
\section{Introducing \wawawriter}
In this section, we describe \wawawriter, a next-generation AI-assisted writing platform we build to fully unleash the capabilities of \baby. \wawawriter integrates key features of recent AI-assisted writing platforms (e.g., Notion AI) including AI-assisted generation, polishment, and summarization while also implementing a few new innovations for next-generation AI-writing experience. We describe these innovations in the following sections.

\subsection{Human-AI Collaborative Writing}
One major innovation in \wawawriter is a new interface for human-AI collaborative writing, which delivers a drastically different user experience compared to traditional AI-assisted writing platforms. Thanks to the Agents~\citep{agents} framework, we are able to build controllable writing agents that act like independent human collaborators/co-authors in standard collaborative editors such as Google Docs or Notion. The writing agents understands the goal of the current document by reading customized settings such as the title or a short description of the document. It then takes actions according to the current content in the document and the recent actions of human users (or other writing agents) that reveal their focus. Human users can also chat with the writing agents in a chat interface to instruct them what to do. The ability of writing agents to use both external APIs such as web search and build-in editor APIs such as bolding or adjusting the line space enables them to accomplish tasks much more complex than what conventional AI assistants can do. With the human-agent interaction feature in the Agents framework, \bench also supports collaborative editing between multiple human writers and language agents. Users can customize their multiple writing agents and collaborate with one or a few of them when writing stories or articles. Users can specify tasks for each writing agent while multiple writing agents can also communicate with each other to autonomously distribute labors.

\subsection{Integration of External Knowledge and Tools}

Another new feature of \wawawriter is that users can now build their own personal knowledge bases via document uploading or saving web pages. \wawawriter automatically organizes and summarizes the knowledge base and then uses them as references when writing stories and articles. Specifically, we prompt an LLM to split documents into chunks based on their semantics, embed them with our embedding model, and store them in a VectorDB. During writing, we dynamically retrieve the entries of the user's personal knowledge base using semantic search using the current context in the user's editor as the query. Following Socratic Models~\citep{socratic}, our knowledge base also supports images in documents by using GPT-4V to do detailed captioning for each image and then using the captions as entries representing the corresponding images. Users can also edit the documents in their personal knowledge bases using all AI-writing features in \wawawriter. In addition, writing agents described in the previous section can also access the personal knowledge base of a user through function calling.

\subsection{Personalized Writing Assistance}
Different from current AI-assisted writing systems, \wawawriter provides personalized writing assistance for different users that suits their writing styles and content preferences. To achieve, we maintain a text-based user profile for each user which describes some basic writing habits and styles (e.g., choice of words and punctuation, preference for the length of sentences, etc.) of the user. The user profile is periodically updated using an LLM according to the recent texts written by the user with a carefully a designed prompt. The user profile is then used as a prefix in the prompt for \baby. In addition to text-based user profiles, we also retrieve paragraphs that are most similar to the current context in the editor and use them as references for RAG. 

\subsection{Infinite Long Text Generation}
\wawawriter also supports infinite long text generation since \baby natively supports the recurrent prompting technique proposed by~\citep{recurrentgpt}. Specifically, to generate a very long text, we iteratively prompt \baby to generate an outline based on the current context and then generate a paragraph of text based on the generated outline. \wawawriter integrates the ``step-by-step'' mode and the ``continuous'' mode in RecurrentGPT, where the next outline is either manually selected by the user or automatically selected by an LLM. As discussed in \citet{recurrentgpt}, this recurrent prompting mechanism drastically improves the creativity, consistency, and relevance of the generated long text, this is especially helpful for story/fiction writing with \wawawriter.

\section{Discussion}
In this technical report, we introduce \baby, a family of LLMs specialized for writing endeavors. \baby is continually pre-trained on carefully curated datasets and then aligned to the preferences of professional writers and editors using a novel data synthesis framework. We also release \bench, the first benchmark for evaluating the writing capabilies of LLMs. \bench covers multiple domains and tasks related to writing. We compare \baby with 10+ popular generalist LLMs and find that \ultra is the current state-of-the-art on the benchmark. Our user study also confirms the superiority of \baby in real-world AI-assisted writing scenarios. The results also confirm the effectiveness of our data synthesis pipeline for training domain-specific LLMs.
\bibliography{main}
\newpage
\appendix

\section{Appendix}

\subsection{Author Contributions}
\textbf{Tiannan Wang} is the core contributor of \baby. Tiannan is responsible for continual pre-training, supervised fine-tuning, and preference optimization. Tiannan is also a main contributor for the data synthesis and the benchmark/evaluation process. \\
\textbf{Jiamin Chen} is a main contributor of \baby. Jiamin is responsible for \bench and is also main contributor for data synthesis and model evaluation process. \\
\textbf{Qingrui Jia} is a main contributor for the data synthesis and supervised fine-tuning stages for fiction writing. Qingrui also contributes to the data synthesis process for non-fiction writing. \\
\textbf{Shuai Wang} is responsible for the application and the deployment of \baby and the prompt engineering for \wawawriter. \\
\textbf{Ruoyu Fang} is a main contributor for the data synthesis process for continual pre-training and supervised fine-tuning. \\
\textbf{Huilin Wang}, \textbf{Chunzhao Xie}, and \textbf{Shengwei Ding} are main contributors for the prompts inside \wawawriter. \\
\textbf{Zhaowei Gao}, \textbf{Chunzhao Xie}, \textbf{Jihong Dai}, \textbf{Jialong Wu}, \textbf{Long Li}, \textbf{Zhiwei Huang} contributed to the data synthesis process for non-fiction writing. \\
\textbf{Chuou Xu}, \textbf{Yibin Liu}, \textbf{Xinle Deng} contributed to the evaluation and benchmarking process. \\
\textbf{Teng Yu}, \textbf{Jiayang Huang}, \textbf{Gangan Ma}, \textbf{Han Xiao}, \textbf{Zixin Chen} \textbf{Gangan Ma},\textbf{Yiru Wang}, \textbf{Siran Ding} are responsible for marketing and operation of \wawawriter and contributed to the product. \\
\textbf{Jiayi Xu}, \textbf{Yilihamu Tayier}, \textbf{Zhenyu Hu}, \textbf{Yuan Gao}, \textbf{Chegnfeng Zheng}, \textbf{Yueshu Ye} are responsible for the implementation of \wawawriter. \\
\textbf{Lei Wan}, \textbf{Siyu Cheng}, \textbf{Xinyue Jiang}, \textbf{Siyu Cheng}, and \textbf{Zhule Song} are responsible for the product design of \baby. \\
\textbf{Xiangru Tang}, \textbf{Xiaohua Xu}, \textbf{Ningyu Zhang}, \textbf{Huajun Chen} are  academic collaborators that contributed to the discussion process and the revision of the technical report. \\ 
\textbf{Yuchen Eleanor Jiang} and \textbf{Wangchunshu Zhou} are project lead and are responsible for the conceptualization, division of labor, and project management for all parts of \baby training, \bench construction, and the productization of \wawawriter. They write the technical report together.

\subsection{Acknowledgments}
We would like to thank Canwen Xu for his insightful discussion, help on revising the draft, and especially for his suggestion on naming the paper. We would also like to thank APUS for their support on computation resources, ABAKA.AI for their support on data collection, and Zhejiang University for general supports.






\subsection{Case Study}

We present a few case studies of content generated by \ultra and GPT-4:
\begin{figure}
    \centering
    \includegraphics[width=\textwidth]{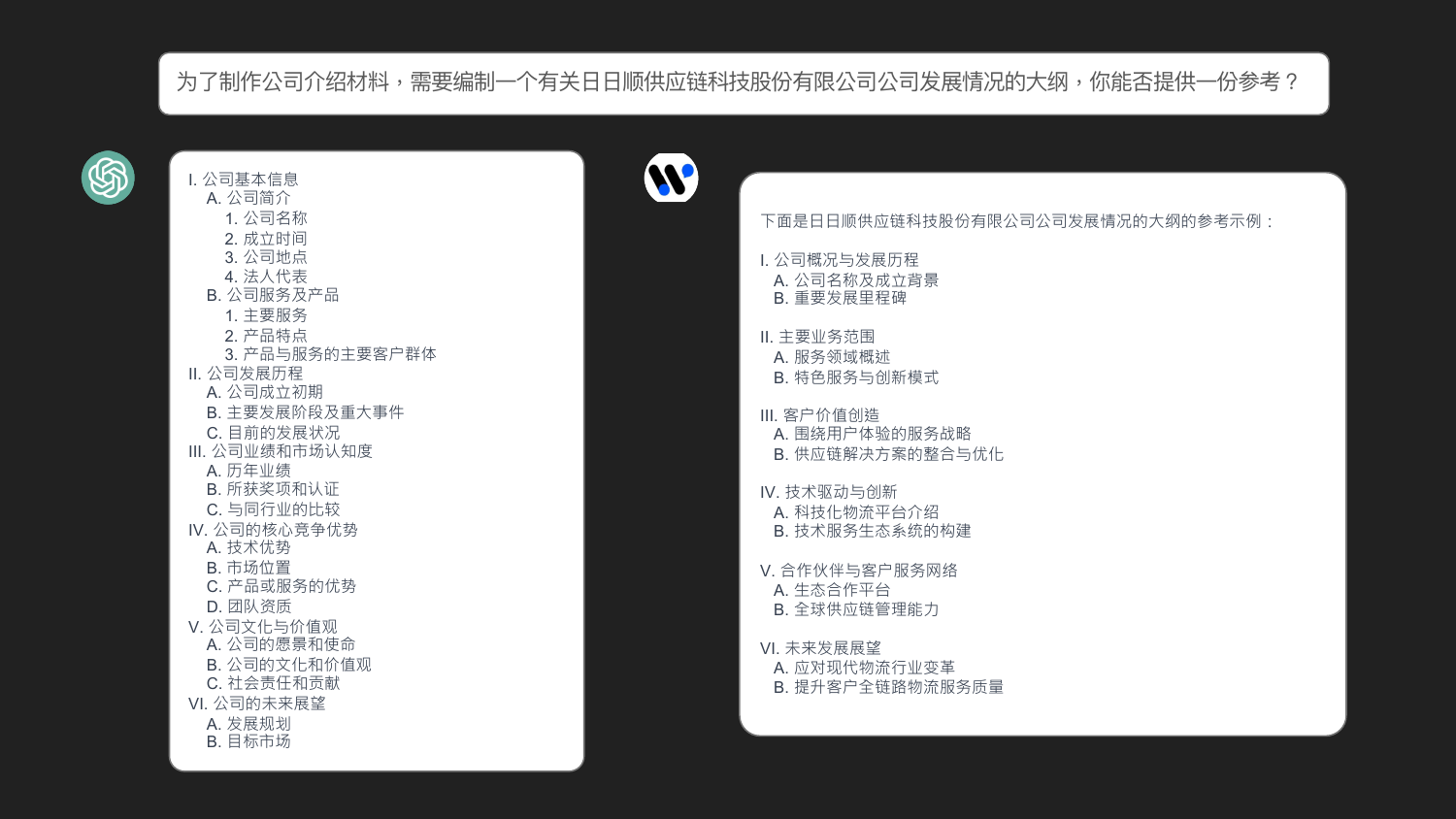}
\end{figure}
\begin{figure}
    \centering
    \includegraphics[width=\textwidth]{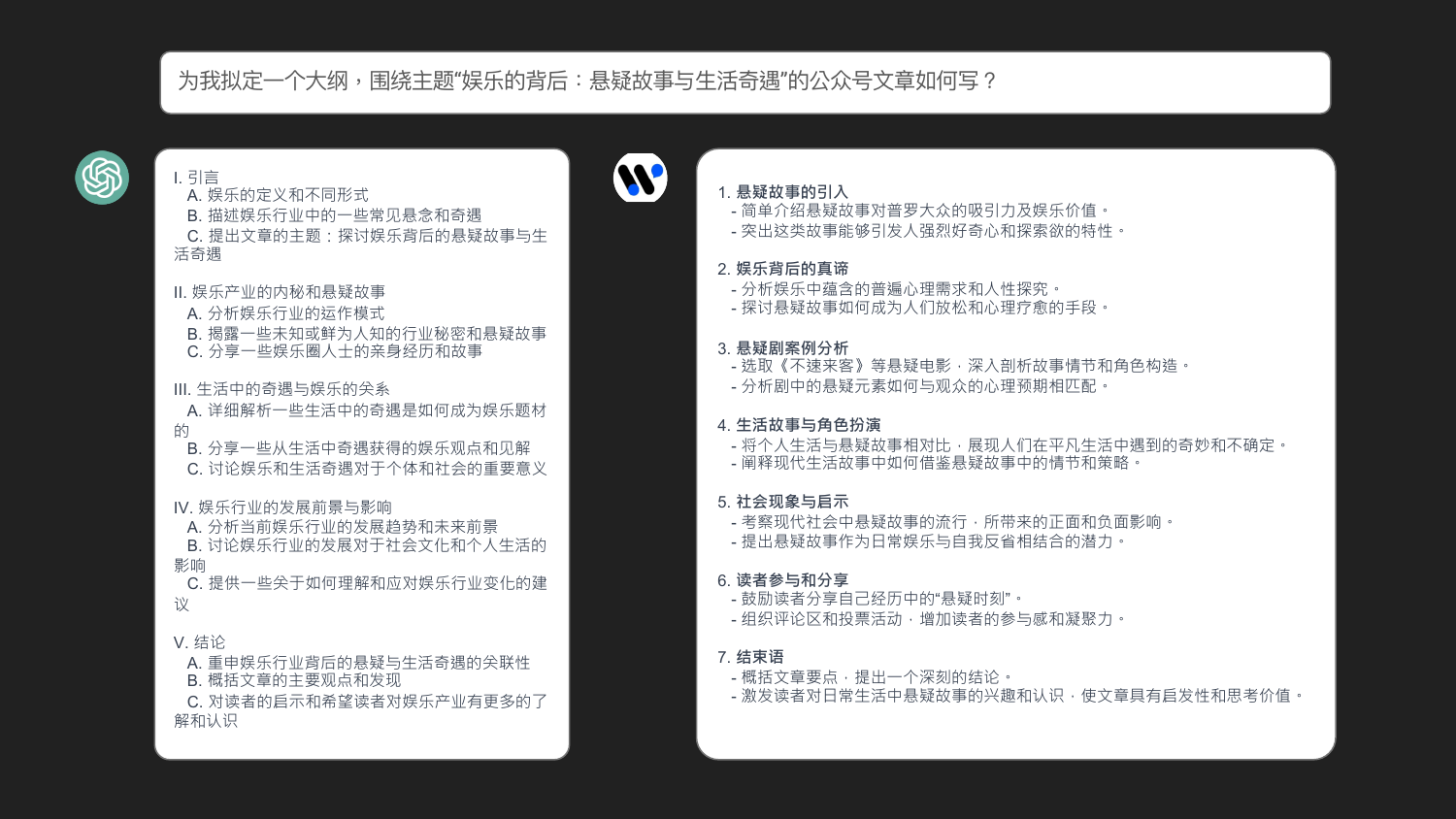}
\end{figure}
\begin{figure}
    \centering
    \includegraphics[width=\textwidth]{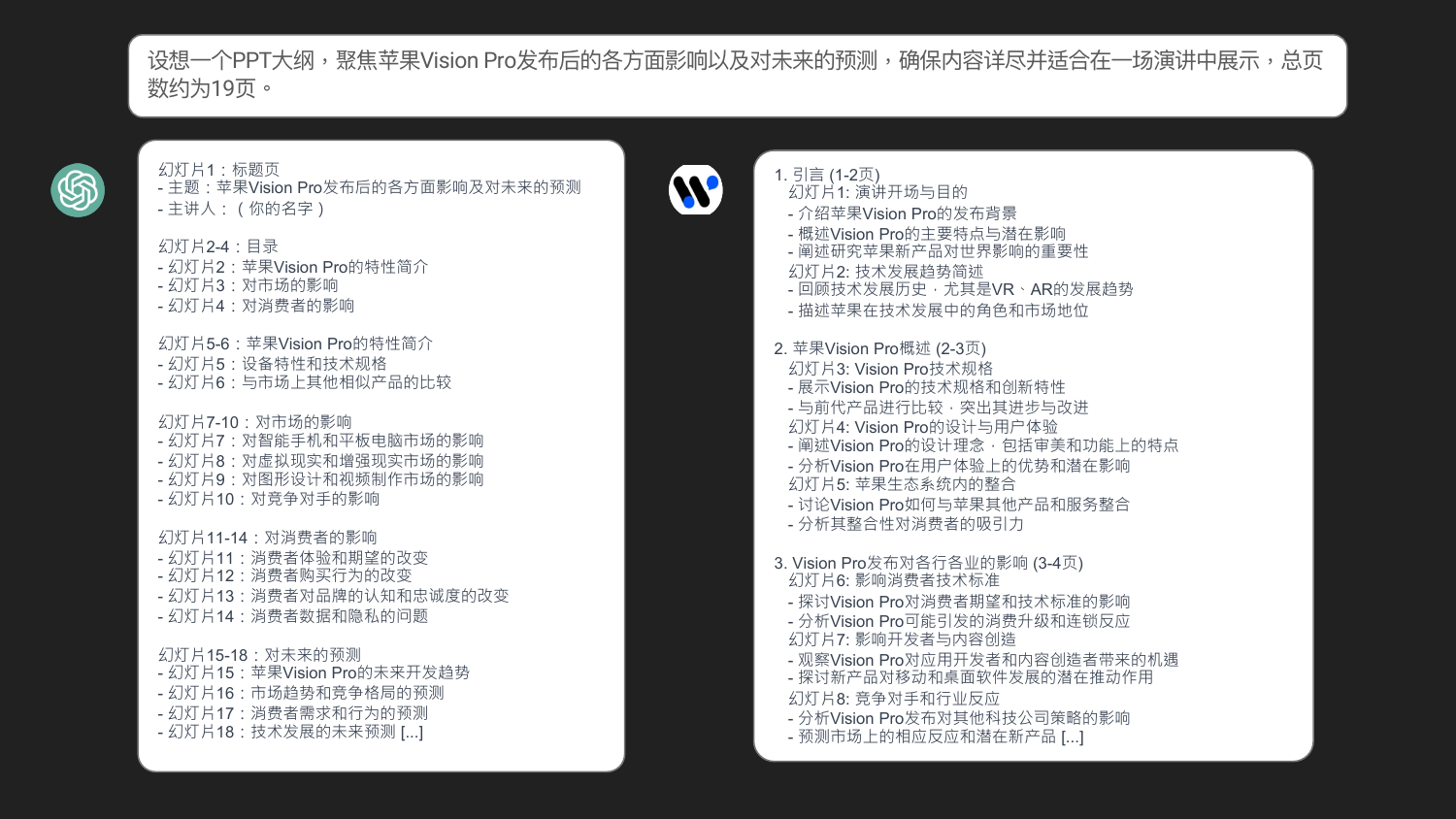}
\end{figure}
\begin{figure}
    \centering
    \includegraphics[width=\textwidth]{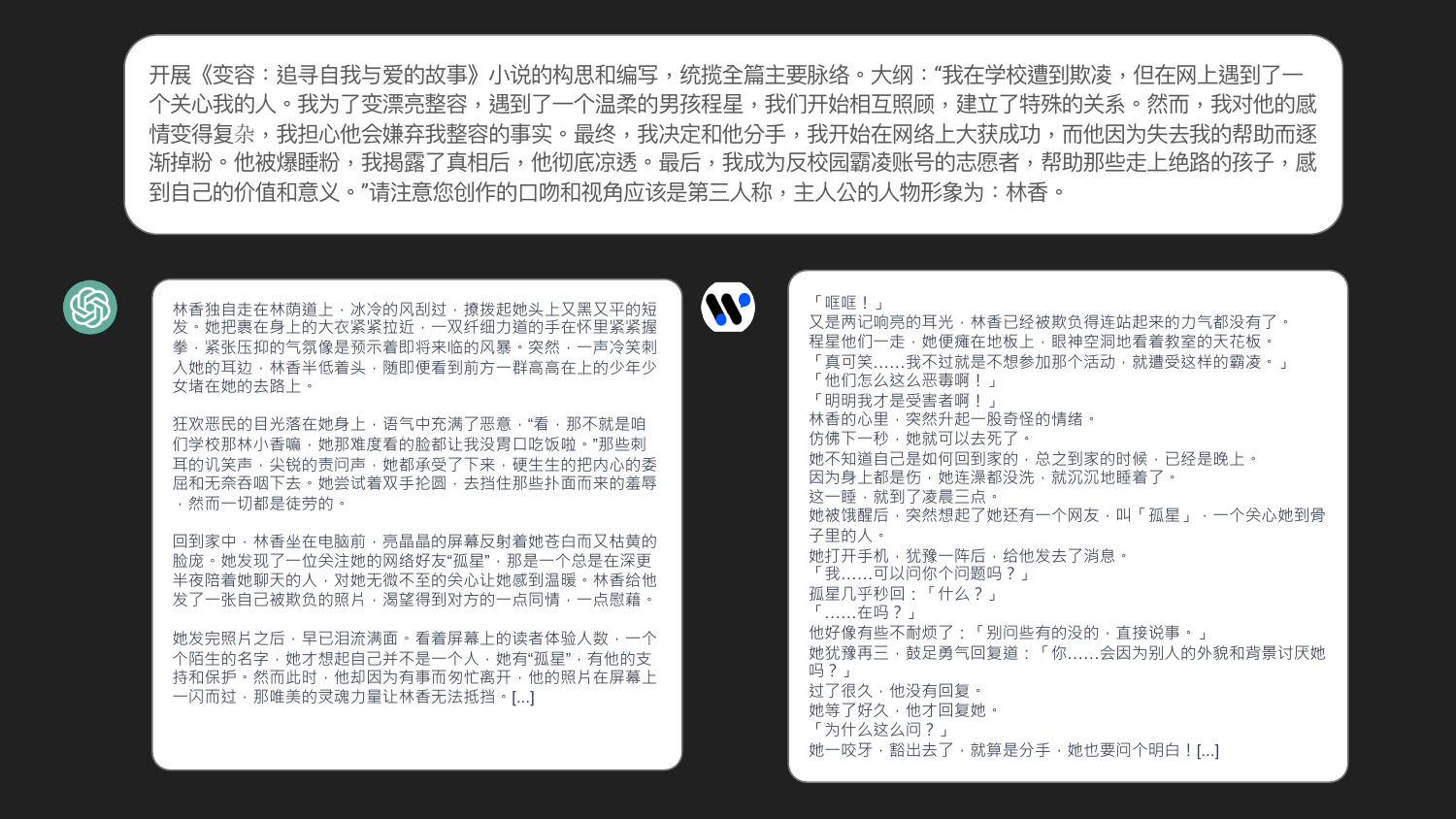}
\end{figure}
\begin{figure}
    \centering
    \includegraphics[width=\textwidth]{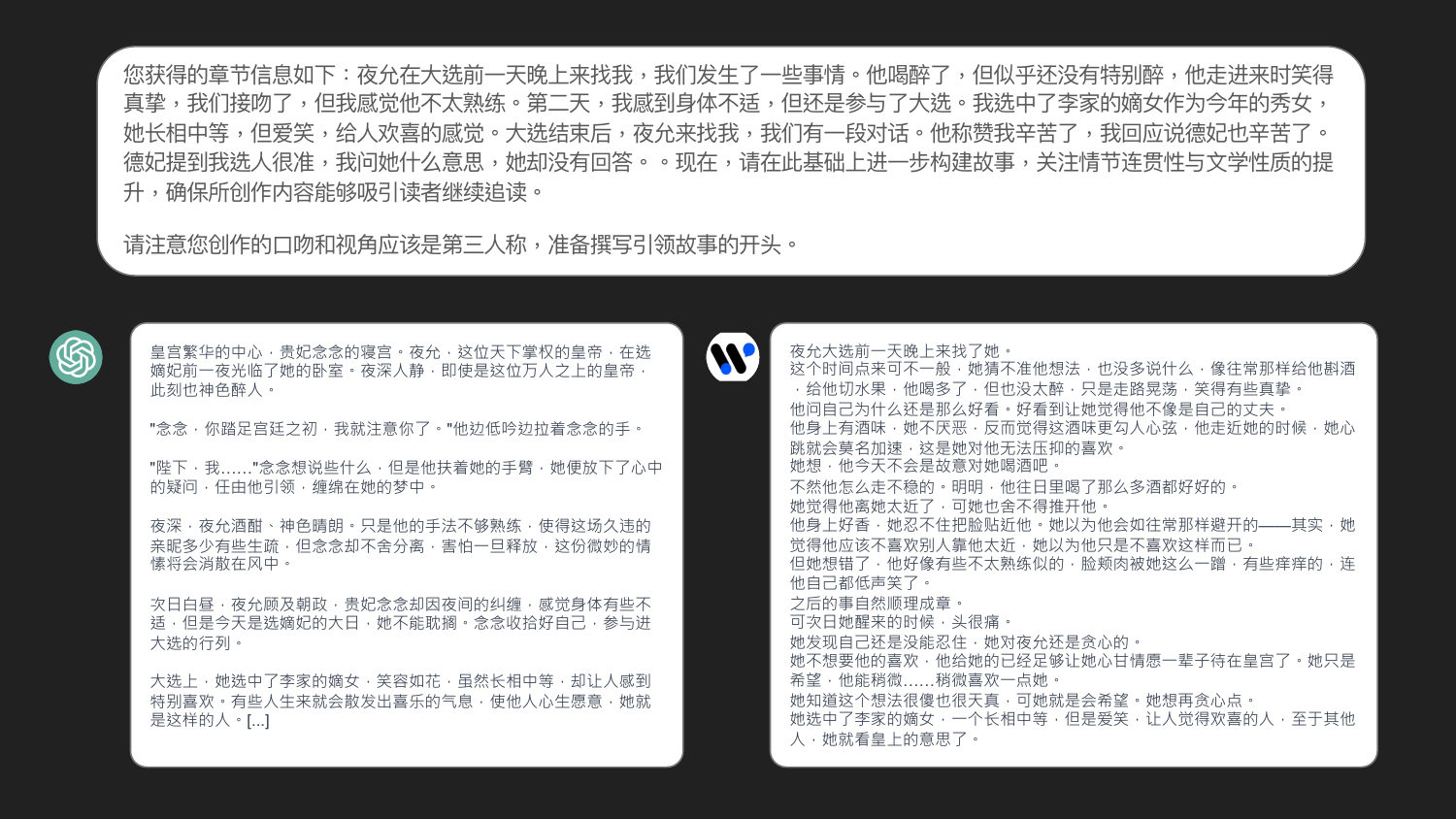}
\end{figure}

\end{CJK*}
\end{document}